\documentclass[runningheads]{llncs}

\usepackage{xspace}

\newcommand{\OUR}{\texttt{GRETEL}\xspace}

\newcommand{\ie}{i.e.\xspace}
\newcommand{\eg}{e.g.\xspace}

\newcommand{\Quote}[1]{\textit{``#1''}\xspace}

\usepackage{amsmath}
\usepackage{amssymb}
\usepackage{graphicx}
\begin{document}
\title{\OUR: A unified framework for Graph Counterfactual Explanation Evaluation\thanks{Research carried out with the help of the HPC \& Big Data Laboratory of DISIM, University of L'Aquila.}}
%
%
\author{
Mario Alfonso Prado-Romero\inst{1} 
\and
Giovanni Stilo\inst{2} 
}
\authorrunning{M.A. Prado-Romero and G. Stilo}
%
\institute{
Gran Sasso Science Institute, 67100 L'Aquila, Italy\\
\email{marioalfonso.prado@gssi.it}\\ \and
Università degli Studi dell'Aquila, 67100 L'Aquila, Italy\\
\email{giovani.stilo@univaq.it}\\
}
\maketitle              
\begin{abstract}
Nowadays, Machine Learning (ML) systems are a fundamental part of those tools with an impact on our daily life in several application domains.
Unfortunately those systems, due to their black-box nature, are hardly adopted in those application domains (e.g. health, finance) where having an understanding of the decision process is of paramount importance. For this reason, explanation methods were developed to give insight into how the ML model has taken a specific decision for a given case/instance. In particular, Graph Counterfactual Explanations (GCE) is one of the possible explanation techniques in the Graph Learning domain.
Those techniques can be useful to discover, for example: i) molecular compounds similar in terms of specific desired properties, or ii) new insights into the interplay of different brain regions for certain diseases.
Unfortunately, the existing works of Graph Counterfactual Explanations diverge mostly in the problem definition, application domain, test data, and evaluation metrics, and most existing works do not compare against other counterfactual explanation techniques present in the literature.
For these reasons, we present \OUR, a unified framework to develop and test GCEs'. Our framework provides a set of well-defined mechanisms to easily integrate and manage: both real and synthetic datasets, ML models, state-of-the-art explanation techniques, and a set of evaluation measures. 
\OUR is a well-organized and highly extensible platform, which promotes the Open Science and experiments reproducibility thus it can be adopted effortlessly by future researchers who want to create and test their new explanation methods by comparing them to existing techniques across several application domains, data and evaluation measures.
To present \OUR, we show the experiments conducted to integrate and test several synthetic and real datasets with several existing explanation techniques and base ML models.

\end{abstract}
\section{Introduction}\label{sec:intro}

Nowadays, Machine Learning (ML) methods are a fundamental part of several tools in different application domains. 
In application domains, like health or finance, having an understanding of the decision process is of paramount importance.
On the opposite, the predictions made by black-box systems, due to their nature, are hardly understandable preventing their wide adoption.
To overcome this limitation explanation methods were developed to give insight into how the ML model has taken a specific decision for a given case/instance \cite{guidotti2018survey}.

Since their creation Graph Neural Networks (GNN) \cite{scarselli2008graph} have attracted the interest of the ML community because they allow leveraging the advantages of Deep Neural Networks (DNN) on graph data. However, this also means that GNNs behave as black boxes. Given the particularities of graph data, many explanation techniques have had to be developed specifically for GNNs.
In particular, Graph Counterfactual Explanations (GCE) is one of the possible explanation types in the Graph Learning domain. A counterfactual explanation answers the question: \Quote{what changes should I do to the input in order to obtain a different output}. GCE techniques can be useful to discover, for example, i) molecular compounds similar in terms of specific desired properties, or ii) new insights into the interplay of different brain regions for certain diseases.
Existing works (presented in section \ref{sec:related}) of Graph Counterfactual Explanations diverge mostly in the problem definition, application domain, test data, and evaluation metrics, and most existing works do not compare against other counterfactual explanation techniques present in the literature, making it difficult to promote advancement of this research field.

For these reasons, we present \OUR (in section \ref{sec:framework}), a unified framework to develop and test GCEs'. Our framework provides a set of well-defined mechanisms to easily integrate and manage: both real and synthetic datasets, ML models, state-of-the-art explanation techniques, and a set of evaluation measures. 
Moreover, \OUR is a well-organized and highly extensible platform, which promotes the Open Science and experiments reproducibility thus it can be adopted effortlessly by future researchers who want to create and test their new explanation methods by comparing them to existing techniques across several application domains, data and evaluation measures.
In section \ref{sec:proof}, we prove GRETEL's flexibility by showing the evaluations conducted to integrate and test several synthetic and real datasets with several existing explanation techniques and base ML models.

\section{Related Works}\label{sec:related}

Accordingly to \cite{guidotti2018survey,bodria2021benchmarking} 
the field of eXplainable Artificial Intelligence (XAI) distinguishes methods that are explainable by design from the black-box ones. The explainable-by-design methods are intrinsically explainable; thus, the reasoning for reaching a decision is directly accessible due to the transparency of the ML model. On the other hand, the explanations of Black-box methods are achieved using Post-Hoc methods that build possible explanations for non-interpretable ML models.
There are two main types of explanations: the Factual and the Counterfactual. Factual explanations explain which features and values of the input instance drove the ML model to take a particular decision. In contrast, the Counterfactual explanations produce a new instance related to the input instance but which exposes a different  ML model decision. 

While there are many works focused on explaining Deep Neural Models on image and text data \cite{simonyan2013deep,selvaraju2017grad,dabkowski2017real}, on the opposite, the explanation techniques focused on GNN\footnote{Graph Neural Network, is a neural network that can directly be applied to graphs. It provides a convenient way for node level, edge level, and graph level prediction tasks.} are not very explored, and their study has started very recently \cite{ying2019gnnexplainer,yuan2020explainability}. GNNs' as most Deep Neural Networks are black-box models and their decisions are commonly explained using post-hoc techniques. Furthermore, most explanation methods for GNNs' focus on providing factual explanations \cite{luo2020parameterized,yuan2021explainability,huang2020graphlime}.

Just a few works have explored Counterfactual Explanations for Graphs so far. When producing a counterfactual example it is important to consider how far is this example from the original instance. The closest counterfactual to the original instance its called a minimal counterfactual instance. Considering the distance between the original instance and the counterfactual is important to avoid useless answers i.e "If you become a billionaire you will get a car loan". Unfortunately, some works do not try to find minimal counterfactual explanations \cite{faber2020contrastive,wu2021counterfactual,zhao2021counterfactual,bajaj2021robust}. At the moment, we decided to focus our framework on the methods that try to produce minimal counterfactual examples. 
Among the works providing minimal counterfactual explanations, \textbf{CF-GNNExplainer} \cite{lucic2022cf} follows a perturbation-based approach using edge deletions. Its loss function encourages the counterfactual instance to be close to the original instance. CF-GNNExplainer is focused on providing explanations for the node classification problem.
On the opposite, \textbf{MEG} \cite{numeroso2021meg} allows deletion and addition actions over the input instance for generating the counterfactual examples. The method is based on a multi-objective Reinforcement Learning problem which allows to easily steer towards the generation of counterfactuals optimizing several properties at a time. MEG is specifically designed to provide counterfactual explanations for molecular graphs. The use of domain knowledge helps to generate better explanations but limits the general applicability of this method.
Another method designed for the molecular domain is \textbf{MACCS} \cite{wellawatte2022model}. The main difference with MEG \cite{numeroso2021meg} is that the method does not comprise the reinforcement learning phase.
Lastly, the \textbf{Bidirectional Counterfactual Search} \cite{abrate2021counterfactual} is a method designed for brain networks. Using edge additions and removals, the authors perturb the original instance to obtain a counterfactual example. The general idea is a two-stage heuristic where in the first stage, the original instance is perturbated until a counterfactual example is found. Then, the second stage tries to reduce the distance between the counterfactual and the original instance. The method assumes all graph instances in the dataset contain the same vertices.

Moreover, the newborn research area of Counterfactual Graph Explanation does not have an established set of measures. Each work is typically tested with a specific domain and dataset without providing an exhaustive comparison with the others works present in the literature.
Those lacks justify the necessity of having an available established evaluation procedure and framework. Thus \OUR represents the first work that can be adopted to conduct reproducible advancements in the research field of Graph Counterfactual Explainability.

\section{\OUR Evaluation Framework}\label{sec:framework}

In this section, we will discuss at a high-level the design principles and the core components of the \OUR Evaluation Framework. While in section \ref{sec:implemented} we will provide a complete list of the implemented and thus available components.

\subsection{Design Principles}
\begin{figure}[ht]
  \centering
  \includegraphics[width=\linewidth]{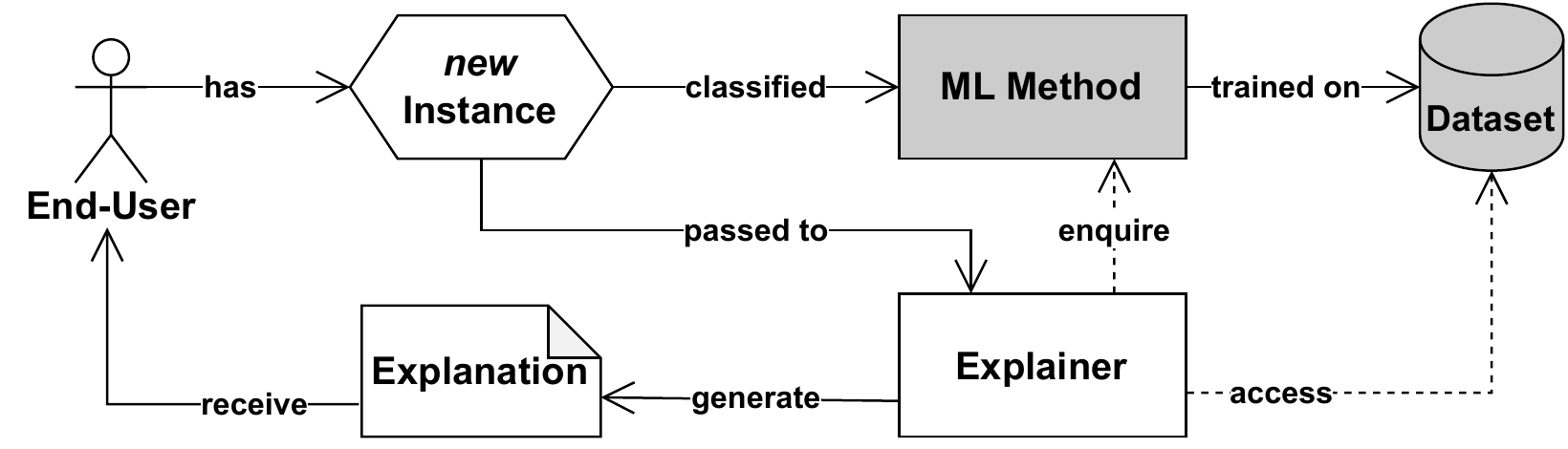}
       \vspace{-2.0em}
  \caption{Typical workflow for the explanation of an instance classified by a black-box model.}
   \label{fig:exai-workflow}
\end{figure}

To better understand the concepts behind the design of the \OUR Framework, we need to provide, as depicted in figure \ref{fig:exai-workflow},  an overview of the typical workflow that is realised when an end-user wants to classify a \Quote{new} instance\footnote{The instance must be considered new because is not part of the original dataset.} comprehensive its explanation.

The overview considers the classification task as a reference, but this task can be substituted by any generic ML task (\ie regression, anomaly detection, etc.).
One assumption that must be considered is that the ML Model is a black-box one and the explanation will be realised through a post-hoc explainability method (from now on, Explainer for simplicity). Thus, as highlighted in grey in figure \ref{fig:exai-workflow}, the model must be already trained on the reference dataset.

Now, supposing that the end-user has a new instance, this will be submitted to the ML Model and to the Explainer to obtain respectively the classification and its explanation. Behind the scene, what typically happens is that the Explainer might enquiry the ML model to understand its behind mechanisms, and/or access the original dataset\footnote{This might also vary by the reference scenario of the task \eg when it must be necessary to preserve privacy, the Explainer might not access the original dataset by default.} to produce the final explanation.
The interactions among the Explainer, the ML Model and the Dataset are highlighted with a dashed line to capture their not mandatory nature that depends on the application scenario and the used Explainer.

Considering the workflow explained above, we need now to clarify which was the point of view and the goals that guided our design process.

The framework was designed keeping in mind the point of view of a researcher who wants to perform an exhaustive set of evaluations. Moreover, the framework must be easily used and extended ( in terms of domain, methods, datasets, metrics and ML models) by future researchers.
We designed \OUR by adopting the \textit{OO}\cite{booch1982object} approach, where the framework's core is constituted mainly by abstract classes which need to be specialised in their implementations.
To promote the framework extendability, we adopted the design pattern \Quote{Factory Method} \cite{lasater2006design}, and we leveraged the configuration files as constituting part of the running framework.

Moreover, to enhance the reproducibility of the evaluations, we not only provide the complete framework\footnote{\OUR is available at \url{https://github.com/MarioTheOne/GRETEL}.} with its configurations, but we also provide and allow storing and loading of the already trained ML Models and the included datasets. Thus, the approach that we followed was to generate or train a dataset or a model on the fly if it was not already stored and readily available. This will mitigate the efforts needed to evaluate new settings (\eg same explainer with other datasets or/and measures).

Lastly,  since the Explainer uses the ML Model agnostically, it then is seen as an Oracle to be enquired. For this reason, in our framework, as typically happens in the XAI domain, we refer to the ML Model as the Oracle.

\subsection{Core Components}\label{sub:core}

\begin{figure}[ht]
 \vspace{-1.0em}
  \centering
  \includegraphics[width=\linewidth]{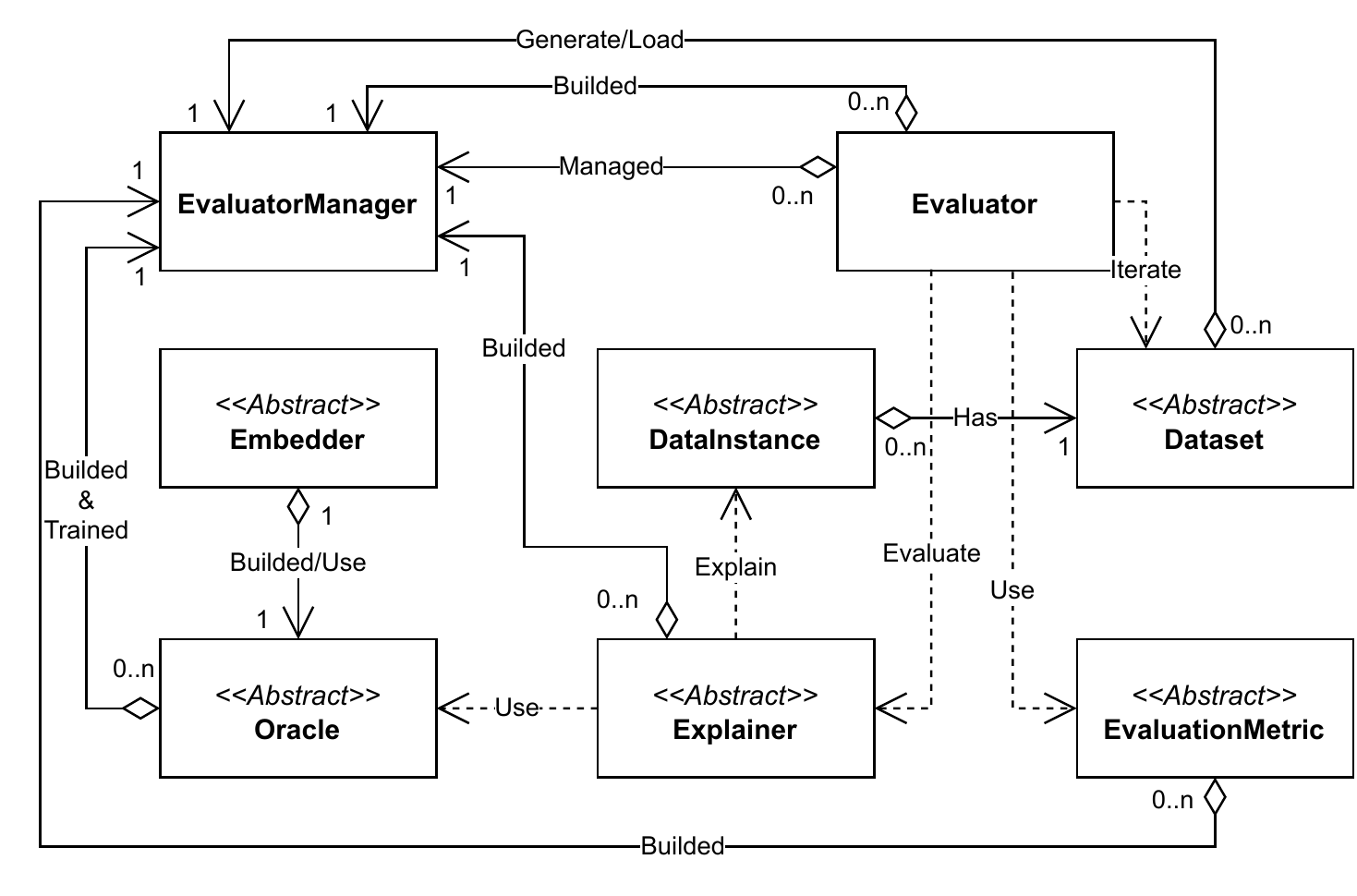}
     \vspace{-2.0em}
  \caption{Overview of the main classes and their relations of GRETEL Evaluation Framework.}
   \label{fig:overview}
\end{figure}

As depicted by the figure \ref{fig:overview}, we discuss the core components of the \OUR Evaluation Framework, providing a brief description of them and their relationships.

Since the framework is focused on the evaluation, the two principal concrete components that we start to discuss are the Evaluator and the EvaluatorManager:

The \textbf{Evaluator} is the component responsible for carrying on the evaluation of a specific Explainer.

The Explainer must be evaluated on a constituted set of Metrics accordingly to one specific Dataset and a reference Oracle. Thus, the Evaluator $e$ can be seen as a tuple $e = <x, d, o, M>$ where $x \in X$ is one of the explainers that must be evaluated, $d \in D$ is the considered Dataset, $o \in O$ is the reference Oracle and $M \subseteq \mathbb{M}$ is the set of Metrics that must be evaluated. Then the Evaluator performs the evaluation by collecting all the performances for each  instance of Dataset $d$.

The aim of the \textbf{EvaluatorManager} is to facilitate the task of running experiments specified by the configuration files and instantiating all the components needed to perform the complete set of evaluations.
Thus, it starts by reading the configuration file that describes all the evaluations and then generates accordingly to it through the factory classes all the different components of the evaluation without the need to use the constructors of the specific subclasses. 
Once the sets of Oracles $O$, Explainers $X$, Datasets $D$ and Metrics $\mathbb{M}$ are instantiated, the EvaluatorManager proceed to create appropriately the Evaluators $E$ that will be responsible for performing the individual evaluations.

The \textbf{DataInstance} class provides an abstract way to interact with data instances. It provides graph representations of the data, class labels and other fundamental information. The DataInstance can be specialised accordingly to the specific necessity that comes from different domains, as it happens for the MolecularDataInstance, which holds specific functionalities to represent the molecular graph in the \Quote{"smiles"} format\cite{weininger1988smiles}.

The \textbf{Dataset} class manages all the details related to generating, reading, writing and transforming the data accordingly to the generating/loading on the fly strategy described before. The Dataset class can be specialised to include specific information, as in the case of the MolecularDataset.

The \textbf{Oracle} class provides a generic interface for interacting with ML models. The main methods exposed by this class are $Embed$, $Fit$ (to data), and $Predict$. The base class also embeds the logic used to keep track of the number of calls that an explainer makes to the oracle, which is an important metric. Each Oracle also provides the specialised mechanisms needed to save and load the trained model on the disk in a way that it can be loaded or trained on the fly. Since some Oracles need to use embedding methods, we also defined the \textbf{Embedder} class which is typically coupled one-to-one with an Oracle.

The \textbf{Explainer} is the base class used by all explanation methods. It exposes the method  $Explain$, which takes an instance and an Oracle as input and returns its explanation. Our goal was to keep this class as simple as possible because it is the one used to encapsulate explanation methods. Moreover, any researcher that wants to test a new explainer in our framework needs to extend it by providing its specific implementation.

The \textbf{EvaluationMetric} is the base class needed to define a specific metric that will be used to evaluate the quality of the Explainer.

\section{Proof of Concept}\label{sec:proof}

Here, we provide the proof of concept for the \OUR Framework by first providing a description of all the components that were implemented accordingly to the core classes (see section \ref{sub:core}) and then by presenting the evaluations conducted with them.
\vspace{-0.9em}
\subsection{Available Implementations}\label{sec:implemented}
\subsubsection{Datasets with statistics}:
 
\textbf{Tree-Cycles} \cite{ying2019gnnexplainer}: Synthetic data set where each instance is a graph. The instance can be either a tree or a tree with several cycle patterns connected to the main graph by one edge. The graph is binarily categorised if it contains a cycle or not. The number of instances, nodes per instance and cycles can be controlled as parameters. The dataset can be generated on the fly and stored, or it can be loaded to promote the reproducibility of the evaluation.

\textbf{ASD} \cite{abrate2021counterfactual}: Autism Spectrum Disorder (ASD) taken from the Autism Brain Imagine Data Exchange (ABIDE) \cite{craddock2013neuro}. Focused on the portion of the dataset containing children below nine years of age \cite{lanciano2020explainable}. These are 49 individuals in the condition group, labelled as Autism Spectrum Disorder (ASD), and 52 individuals in the control group, labelled as Typically Developed (TD).

\textbf{BBBP} \cite{wellawatte2022model}: Blood-Brain Barrier Permeation is a molecular dataset. Predicting if a molecule can permeate the blood-brain barrier is a classic problem in computational chemistry. The most used dataset comes from Martins et al.\cite{martins2012bayesian}. It is a binary classification problem with molecular structure as the features.

\begin{table}[ht]
\vspace{-1.0em}
    \centering
    \begin{tabular}{|l|r|r|r|r|r|r|r|}
    \hline
    dataset &  \#inst &  $|V|$ & $\sigma(|V|)$ & $|E|$ & $\sigma(|E|)$ &  $|C_0|$ &  $|C_1|$ \\
    \hline
    Tree-Cycles & 500 &  300 & 0 & 306.95 &  12.48 & 247 & 253 \\
    ASD &  101 &  116 &   0 &   655.62 &   7.29 & 52 & 49 \\
    BBBP & 2039 & 24.06 & 10.58 & 25.95 & 11.71 & 479 & 1560 \\
    \hline
    \end{tabular}
 \caption{For each dataset it is presented the number of instances, the mean and the standard deviation of the number of vertices ($|V|$, $\sigma(|V|)$) and edges ($|E|$, $\sigma(|E|)$), and the number of instances for each class $|C_0|$ and  $|C_1|$.}
\label{tab:datasets}
\vspace{-2.0em}
\end{table}

\vspace{-0.8em}
\subsubsection{Oracles}:

\textbf{SVM} \cite{cortes1995support}: Support Vector Machine Classifier is a very popular ML model. This oracle also requires the use of an embedder.  Thus Graph2Vec \cite{narayanan2017graph2vec} is available as an embedder.

\textbf{ASD Custom Oracle}: This oracle is provided by Abrate et al. \cite{abrate2021counterfactual}, and it is specific for the ASD dataset. The oracle is based just on some simple rules but can perform better than other oracles on this dataset, given the low number of training instances.

\textbf{GCN} \cite{kipf2017semi}: Graph Convolutional Network is an ML model that can work directly with the graph matrices without previously using an embedder. This particular implementation also considers node types besides the network structure.

\vspace{-1.0em}
\subsubsection{Explainers}:

\textbf{DCE Search}: Distribution Compliant Explanation Search,  mainly used as a baseline, does not make any assumption about the underlying dataset and searches for a counterfactual instance in it.

\textbf{Oblivious Bidirectional Search (OBS)} \cite{abrate2021counterfactual}: It is an heuristic explanation method that uses a 2-stage approach. In the first stage, changes the original instance until its class changes. Then, in the second stage, reverts some of the changes while keeping the different class. This method was developed for Brain Graphs, so it assumes that all graphs in the dataset contain the same nodes.

\textbf{Data-Driven Bidirectional Search (DBS)} \cite{abrate2021counterfactual}: It follows the same logic as OBS. The main difference is that this method uses the probability (computed on the original dataset) of each edge to appear in a graph of a certain class to drive the counterfactual search process. The method usually performs fewer calls than OBS to the oracle while keeping a similar performance.

\textbf{MACCS} \cite{wellawatte2022model}: Model Agnostic Counterfactual Compounds with STONED (MACCS) is specifically designed to work with molecules. This method always generates valid molecules as explanations and thus has the limitation that it cannot be applied to the other domains.

\vspace{-1.0em}
\subsubsection{Evaluation Metrics}\label{sub:metrics}:

\textbf{Runtime (t)}: Measures the seconds taken by the explainer to produce the counterfactual example.

\textbf{Graph Edit Distance (GED) \cite{sanfeliu1983distance}}: It measures the structural distance between the original graph and the counterfactual one.

\textbf{Calls to the Oracle (\#Calls) \cite{abrate2021counterfactual}}:
Given the explainers are model-agnostic and the computational cost of making a prediction sometimes can be unknown, it is a desirable property that the explainers perform as few calls to the oracle as possible.

\textbf{Correctness (C)}: We introduced this metric to control if the explainer was able to produce a valid explanation (\ie the example has a different classification). Given the original instance $G$, the instance $G'$ produced by the explainer and the machine learning model $\Phi$ then correctness returns $1$ if $\Phi(G) \neq \Phi(G')$ and $0$ if $\Phi(G) = \Phi(G')$

\textbf{Sparsity (S)} \cite{yuan2020explainability}: Be $|G|$ the number of features in the instance, and $D(G,G')$ the edit distance between the original instance and the counterfactual one, then we can define Sparsity as:
        \begin{equation}
           Sparsity = 1 - \frac{D(G,G')}{|G|}
        \end{equation}

\textbf{Fidelity (F)} \cite{yuan2020explainability}: It measures how much the explanations are faithful to the Oracle, considering its correctness. Be $G$ the original instance, $y_G$ the ground truth label of the original instance, $G'$ the instance produced by the explanation method, $\Phi$ the ML model, and $I:\mathbf{N} \times \mathbf{N} \longrightarrow \{0,1\}$ a function that returns $1$ if the two labels are the same and $0$ otherwise, then Fidelity is defined as:
        \begin{equation}
            Fidelity = I(\Phi(G), y_G) - I(\Phi(G'), y_G)
        \end{equation}

\textbf{Oracle Accuracy (Acc)} \cite{fawcett2006introduction}: The performance of the oracle can significantly affect the obtained explanations. For this reason it is important to understand how is performing the oracle before evaluating the explainer. Given the original instance $G$, an oracle $\Phi$, and a ground truth label $y_G$ then oracle accuracy returns $1$ if $\Phi(G) = y_G$ and $0$ in other case.

\subsection{Conducted Evaluations}

To prove GRETEL's ability to evaluate explanations across diverse domains, we evaluated the explainers on a synthetic dataset (Tree-Cycles), a brain one (ASD) and a molecular one (BBBP). Table \ref{tab:datasets} reports the general statistics of the included datasets.

We tested three explainers DCE, OBS and DBS, on the Tree-Cycles dataset (see Table \ref{tab:tree_cycles}). Here, none of the explainers outperformed the others consistently. However, we should highlight that thanks to the flexibility of our framework, we were able to evaluate OBS and DBS not only on their native brain dataset for the first time.

\begin{table}[h]
    \centering
    \begin{tabular}{|l|r|r|r|r|r|r|r|}
    \hline
    $Exp.$ & $t (s)$ & $GED$ & $\#Calls$ & $C$ & $S$ & $F$ & $Acc$ \\
    \hline
    DCE & 7.46   & 571.88 & 501    & 1    & 0.63 & 0.86 & 0.93 \\
    OBS & 7.07   & 570.04 & 158.23 & 0.99 & 0.63 & 0.86 & 0.93 \\
    DBS & 302.92 & 581.20 & 812.34 & 0.99 & 0.64 & 0.86 & 0.93 \\
    \hline
    \end{tabular}
    \caption{Explainers Evaluation on the Tree-Cycles dataset using the SVM oracle using the metrics described in \ref{sub:metrics}}
    \label{tab:tree_cycles}
\end{table}

Table \ref{tab:asd} show the conducted evaluations using the ASD dataset. Here, the OBS and DBS outperform the GED of the DCE baseline. Furthermore, these results are comparable to those obtained by Abrate et al. \cite{abrate2021counterfactual}, showing that the integration into our framework did not affect the explanation quality. However, it is notably that DBS performed more calls to the oracle than OBS, contrary to original reports.

\begin{table}[h]
    \centering
    \begin{tabular}{|l|r|r|r|r|r|r|r|}
    \hline
    $Exp.$ & $t (s)$ & $GED$ &  $\#Calls$ & $C$ & $S$ & $F$ &  $Acc$ \\
    \hline
    DCE & 0.09  & 1011.69 & 102    & 1 & 1.31 & 0.54 & 0.7722 \\
    OBS & 3.23  & 9.88    & 340.73 & 1 & 0.01 & 0.54 & 0.7722 \\
    DBS & 83.45 & 11.78   & 362.05 & 1 & 0.02 & 0.54 & 0.7722 \\
    \hline
    \end{tabular}
    \caption{Explainers Evaluation on the ASD dataset using the ASD custom oracle using the metrics described in \ref{sub:metrics}}
    \label{tab:asd}
\end{table}

Using the BBBP dataset we evaluate MACCS against the DCE (see Table \ref{tab:bbbp}). Both methods obtained similar results in the mean of the GED. However, it must be noticed that the correctness of MACCS is much lower due to a bug in the original code. MACCS is less time consuming and makes fewer calls to the oracle than DCE. However, the good results obtained by DCE are notable, considering it does not leverage any domain knowledge.

\begin{table}[h]
    \centering
    \begin{tabular}{|l|r|r|r|r|r|r|r|}
    \hline
    $Exp.$ & $t (s)$ & $GED$ & $\#Calls$ & $C$ & $S$ & $F$ & $Acc$ \\
    \hline
    DCE   & 35.35 & 27.94 & 2040    & 0.99 & 0.59 & 0.72 & 0.86 \\
    MACCS & 31.35 & 11.23 & 1221.33 & 0.40 & 0.19 & 0.23 & 0.86 \\
    \hline
    \end{tabular}
    \caption{Explainers Evaluation on the BBBP dataset using the GCN oracle using the metrics described in \ref{sub:metrics}}
    \label{tab:bbbp}
\end{table}

\section{Conclusions}\label{sec:conclusions}

We presented GRETEL\footnote{\OUR is available at \url{https://github.com/MarioTheOne/GRETEL}}, a unified framework for evaluating and developing Graph Counterfactual Explanation Methods. We first discussed its general architecture with its core components.

To prove its flexibility, we presented the implementations of some datasets (synthetic and real), oracles, and explainers from the literature. Furthermore, we evaluated the implemented explainers in several settings (datasets and oracles), showing that all the methods achieved results similar to those reported in their reference work.

Furthermore, using GRETEL, for the first time, we could compare these methods with the  DCE baseline and test them using datasets coming from other domains. Thanks to the included set of metrics, we have compared Explainers' performances precisely. This shows GRETEL's high potential to help developers test their explanation techniques across diverse domains, datasets, and ML models.

In the future, we would like to: i) enlarge the set of the explanation methods, including those explainers that are trained on the dataset; ii) enable parallel computation in the EvaluatorManager and other modules of the framework to speed up the evaluation time; iii) expose the control over the oracles hyperparameters through the configurations of the framework.

%
%
%
\bibliographystyle{splncs04}
\bibliography{GRETEL}

\end{document}